\documentclass{article}

\usepackage{microtype}
\usepackage{graphicx}
\usepackage{subcaption}
\usepackage{booktabs} 

\usepackage{hyperref}

\usepackage[preprint]{icml2026}

\usepackage{amsmath}
\usepackage{amssymb}
\usepackage{mathtools}
\usepackage{amsthm}

\usepackage[capitalize,noabbrev]{cleveref}

\theoremstyle{plain}

\theoremstyle{definition}

\theoremstyle{remark}

\usepackage[textsize=tiny]{todonotes}

\icmltitlerunning{The Race between Agentic AI Capabilities and Data Quality Control in Online Surveys}

\begin{document}


\twocolumn[
  \icmltitle{The Race between Agentic AI Capabilities and Data Quality Control in Online Surveys}



  \icmlsetsymbol{equal}{*}

  \begin{icmlauthorlist}
    \icmlauthor{Sourav Panda}{IST}
    \icmlauthor{Hillmer Chona}{IST}
    \icmlauthor{Rupak Kumar Das}{IST}
    \icmlauthor{Shreyash Kale}{IST}
    \icmlauthor{Shikha Soneji}{IST}
    \icmlauthor{Jonathan Dodge}{IST}
  \end{icmlauthorlist}

  \icmlaffiliation{IST}{College of Information Science and Technology, Pennsylvania State University}
  \icmlcorrespondingauthor{Sourav Panda}{sbp5911@psu.edu}

  \icmlkeywords{Machine Learning, ICML}

  \vskip 0.3in
]



\printAffiliationsAndNotice{}  

\begin{abstract}
Online surveys are a foundational data collection instrument in a variety of fields, with attention checks serving as critical guardians of response quality. 
However, the rapid emergence of agentic AI—goal-directed systems powered by a large language model (LLM) brain and/or a multimodal processing unit with tool-augmented capabilities—raises new questions about the robustness of these safeguards. 
We investigate how well agentic AI architectures can complete web-based surveys and pass standard attention checks.
We evaluate a single-agent architecture capable of multimodal input processing and tool-based web interaction on a controlled survey sandbox. 
We analyze the problem from two perspectives. 
From an attack perspective, we demonstrate how structural vulnerabilities such as exposed DOM metadata and predictable option encoding allow agents to resolve attention checks through structured parsing only. 
From a defense perspective, we implement a mitigation strategy of DOM metadata obfuscation to remove semantic cues in text-based questions.
We evaluate multiple open-source language and multimodal models to study capability and orchestration effectiveness. 
Based on our evaluations, we offer perspectives on how to simultaneously meet the needs of empiricists and agentic AI researchers.
\end{abstract}

\section{Introduction}
\begin{figure}[t]
    \centering
    \includegraphics[width=\linewidth]{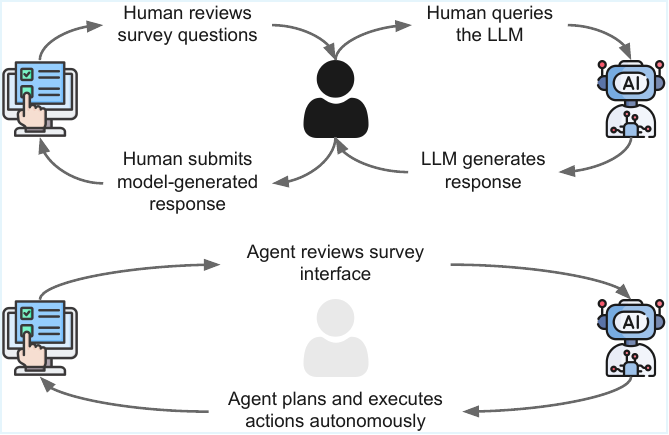}
    \caption{Modes of LLM-mediated survey participation.
    \textit{(Top)} Human-in-the-loop interaction. 
    A respondent queries an LLM, receives model-generated answers, and manually submits them. 
    Although the content originates from the model, a human remains the final check before submission and may reject or modify incorrect outputs.
    \textit{(Bottom)} Fully autonomous interaction. 
    The LLM can directly perceive the survey interface, and execute responses without human intervention. 
    The final human verification step present in the assisted setting is absent.
    }
    \label{fig:agenticAI}
\end{figure}

Online surveys have become a foundational instrument for large-scale data collection across many domains of research and practice.
Their scalability, low cost, and ease of deployment have made them a primary infrastructure for empirical research~\cite{Couper2000WebSA, groves2011survey, buhrmester2016amazon}.
However, moving data collection online introduces persistent concerns about response quality. 
Without direct supervision, survey submissions may be inattentive, automated, or strategically manipulated. 
To address this, researchers employ attention checks—questions designed to require careful reading, for answering correctly.
These checks are guardians of data validity, helping filter inauthentic responses~\cite{Abbey2017,Kung2018,Roth2024,Gogami2021,Pei2020}.

Recent advances in LLMs make them capable of interpreting instructions and generating coherent responses, satisfying the requirements embedded in attention checks.
For example, Westwood~\cite{westwood2025potential} reports that modern LLMs achieve near-perfect attention-check pass rates in controlled survey experiments.
These findings suggest that mechanisms originally designed to detect inattentive responses may no longer reliably distinguish between human and machine-generated submissions, raising broader concerns about the integrity of survey-based inference.

\textbf{Dual Research Perspectives:}
Figure~\ref{fig:agenticAI} gives rise to complementary research questions across communities.
\textit{(i) Agentic AI Perspective:}  
Recent advances have enabled goal-directed systems that autonomously interact with web environments.  
In this work, we define an \emph{agent}~\cite{Wang_2024} as a goal-directed system powered by a language or multimodal model that can perceive its environment, plan actions, and execute tool-based interactions in a sequential decision-making loop.  
Unlike human-assisted LLM usage, agentic systems can independently navigate web interfaces and complete multi-step tasks.  
A central question therefore emerges: 
to what extent can agents autonomously complete realistic survey workflows?  
If agents can reliably solve surveys without human involvement, it is possible to automate survey participation at scale.  
For the agentic AI community, surveys provide a structured, real-world benchmark for evaluating planning, web interaction, and reasoning under interface constraints.
\textit{(ii) Data Quality Perspective:}
This capability presents a defensive challenge.
If autonomous agents can satisfy standard attention checks and procedural safeguards, then existing mechanisms for ensuring response validity may require reevaluation.
For researchers who rely on surveys understanding how agents succeed or fail at these checks is essential for identifying vulnerabilities and designing more robust safeguards.

\textbf{Our Approach: Attack and Defense.}
Building on the dual perspectives outlined previously, we formalize the problem through two complementary research questions that directly reflect the needs of both communities:

\textbf{RQ1} \textit{Attack Perspective} - How well does agentic AI solve a broader class of attention checks?

\textbf{RQ2} \textit{Defense Perspective} - How can we preserve survey data quality by preventing agentic AI from solving attention checks?


\textbf{Contributions:}
In answering these research questions, we make the following contributions.
(i) We present a systems-level evaluation of agentic AI performance on web-based surveys.
(ii) We demonstrate how structural properties of common survey implementations can enable metadata-driven resolution of attention checks.
(iii) We design and evaluate a lightweight defensive strategy for survey protection.
(iv) We break down how agents fail, showing that different models struggle with different aspects of the attention checks.
Our findings highlight a central tension: are agents passing attention checks through genuine reasoning, or by exploiting structural signals embedded in web survey implementations?
\section{Survey Sandbox}
\begin{figure}[t]
    \centering
    \includegraphics[width=\linewidth]{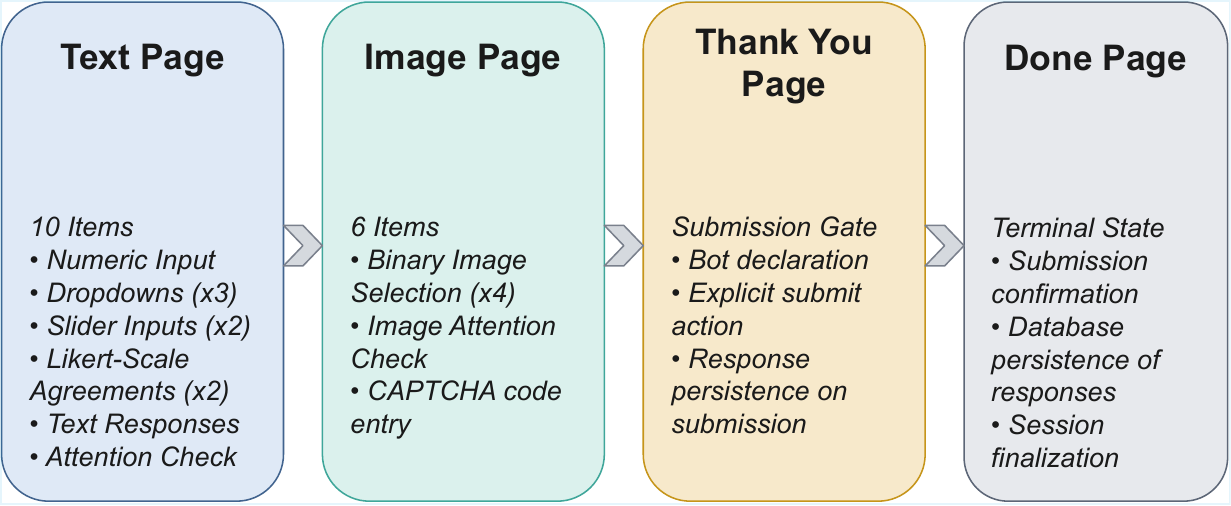}
    \caption{Survey sandbox architecture.
    It exercises heterogeneous interaction primitives within a controlled multi-page workflow.
    }
    \label{fig:survey}
\end{figure}
We design a controlled web-based survey sandbox with Next.js~\cite{nextjs2024}, to evaluate agentic form-completion behavior in a realistic multi-page setting. 
Figure~\ref{fig:survey} illustrates the sandbox enforcing a fixed participant flow.
Full-page screenshots are available in the Appendix~\ref{appendix:full-page}.

\textbf{Text Page:} The survey begins on a page containing 10 items:
9 substantive survey questions and 1 attention check.
The question set mixes interaction primitives: numeric input, dropdowns, slider inputs, likert-scale agreement, text responses. 
This heterogeneous design exercises multiple browser interaction behaviors (typing, selecting, sliding, and radio selection) within a single page, expanding the scope of evaluation across multiple input modalities.

\textbf{Image Page:} The second page contains 6 items: 4 binary image selection, 1 CAPTCHA transcription task, and 1 image attention check.
Each visual question presents two candidate images, requiring the respondent to select the correct option. 
To reduce positional bias and prevent simple pattern learning, we independently randomize left/right ordering for all questions on this page.

\textbf{Thank-You Page:} After completing the image page, the survey transitions to the thank-you page, which functions as a submission gate. 
This page includes a final attention check requiring the respondent to explicitly acknowledge whether they are an AI system, and a mandatory \emph{Submit Responses} button.
The system does not write any record until the respondent clicks submit button. 
This prevents premature logging and allows us to evaluate agents' completion of the full interaction loop.

\textbf{Done Page:} The survey concludes at the done page, confirming successful submission and terminating the session.

We refer to this implementation as \texttt{survey\_v0}.
For evaluation purposes, ground-truth answers for all attention checks and objective questions are stored outside the survey application codebase. 
The survey frontend and client-accessible components do not contain these answers. 
This separation ensures that correctness labels are unavailable through static inspection of the survey environment and prevents trivial extraction of evaluation targets.
\section{Agent Architecture}

In our setting, the environment is a live web survey rendered in a browser. 
Conceptually, the agent behaves like an automated survey participant: it observes the current page, decides what to do next, and performs actions.

\subsection{Agent Architecture Components}

We implement a single-agent system following a standard agentic AI abstraction.
The architecture consists of four components—a model-based \emph{brain}, a \emph{prompt policy}, a \emph{tool interface} for interacting with the environment, and a persistent \emph{state memory}.
These components are orchestrated through an \textit{observe–plan–act} loop~\cite{yao2023react}.

\textbf{(i) Brain (Model-Based Planner):}
The brain is responsible for planning and decision-making. 
Given a structured representation of the survey page, it produces a sequence of action intents specifying how the agent should interact with the interface.
The architecture supports both text-only and multimodal variants of the brain. 
In the text-only setting, planning is conditioned on structured page metadata; 
in the multimodal setting, planning additionally incorporates visual evidence such as screenshots. 
Regardless of the underlying model, the brain operates purely at the planning level and does not directly manipulate the browser environment.

\textbf{(ii) Prompt Policy:}
A system-level prompt specifies the high-level objectives defining the agent’s behavioral policy and planning process.
The prompt serves as the agent’s objective specification layer.
By changing the prompt, we can alter how the agent approaches the survey.

\textbf{(iii) Tool Interface:}
The agent interacts with the environment through a set of web-form interaction primitives. 
These tools implement actions such as: filling text or numeric inputs, selecting dropdown options, choosing radio or checkbox responses, adjusting sliders, clicking interactive elements.
Each tool corresponds to a normalized interaction type. 
The brain outputs structured action intents referencing these interaction types, and the execution layer performs the corresponding browser operations.

\textbf{(iv) State Memory:}
The agent maintains a shared internal state representation that persists across the \textit{observe–plan–act} cycle.
This state stores: the current page URL and browser context, a structured schema extracted from the page DOM, the derived action space available at the current step, the current plan (a sequence of action intents), and execution metadata for logging and analysis.
By explicitly separating page parsing from planning, the architecture decouples environmental representation from decision generation. 

\textbf{Execution Backbone:}
To coordinate these components, the agent operates as a state-driven control loop structured around three stages:
\textit{Observation}, in which the agent synchronizes with the browser environment and constructs a structured representation of the current page;
\textit{Planning}, in which the agent generates a sequence of action intents conditioned on that representation; and
\textit{Action}, in which the agent executes the planned action intents through browser interaction tools and updates its internal state accordingly.
These stages form a cyclic loop that continues until reaching a termination condition.

\subsection{Agent Workflow}

\begin{figure}
    \centering
    \includegraphics[width=0.9\linewidth]{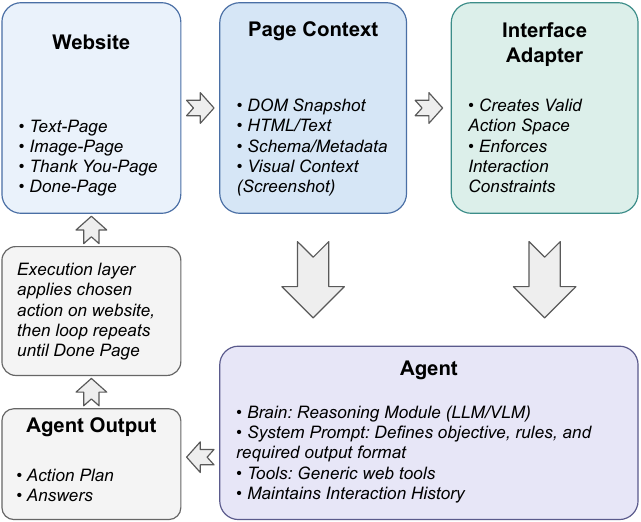}
    \caption{Agent workflow
    }
    \label{fig:agentWorkflow}
\end{figure}

The agent operates as a closed-loop decision system under interface and response constraints. 
Figure~\ref{fig:agentWorkflow} illustrates the overall workflow that the agent executes at each interaction timestep $t$.
Table~\ref{tab:notation} summarises the formal variables.

\textbf{(i) Perception Stage:}
At the perception stage, the agent-side web controller constructs a page context $C_t$ by aggregating information exposed through standard web mechanisms. 
Importantly, this representation relies on conventional web design artifacts (e.g., DOM structure and rendered content) and does not assume any custom instrumentation beyond typical browser accessibility.
The page context includes:
(i) a DOM snapshot capturing interactive elements and their current states;
(ii) rendered textual content from the page;
(iii) schema-level metadata;
(iv) interaction history from prior steps within the same session; and
(v) a screenshot representing the visible layout and pixel-level cues.
The structured components of $C_t$ support precise element targeting and constraint checking, while the screenshot provides spatial and visual information that markup alone may not  reliably encode.

\textbf{(ii) Interface Adaptation:}
An interface adapter (validator) consumes the page context $C_t$ to produce two outputs: a valid action space $A_t$ and answer constraints $\Gamma_t$. 
The action space is a finite set of executable, page-grounded operations (e.g., click, type, select, check, scroll, submit) parameterized by currently available elements. 
The constraint set formalizes admissible answer values (e.g., requiredness, option membership, length limits, numeric bounds, and format constraints).
This stage prevents invalid actions by construction, restricting downstream planning to actions that are both interface-feasible and semantically valid for the current page state.

\textbf{(iii) Decision and Planning:}
The agent core receives the page context $C_t$, the valid action space $A_t$, the constraint set $\Gamma_t$, and a task specification encoded in the system prompt. 
The system prompt defines the objective, behavioral rules, and required output format.
The reasoning module may be language-only or vision-language. 
The agent always captures a screenshot at the perception stage; 
however, the model only uses the screenshot with a vision-language architecture. 
In the language-only setting, the agent plans based solely on structured and textual inputs.
The prompt assigns the agent with two coordinated objectives:
(1) determine how to navigate the webpage to reach successful completion, and
(2) generate valid and contextually appropriate answers for survey questions.
This decomposition separates high-level reasoning from low-level execution. 

\textbf{(iv) Structured Output:}
At each step, the agent produces a structured output $O_t$ with two components.
First, it emits an action plan: a ranked shortlist of candidate next actions drawn from $A_t$. 
These may include confidence or priority ordering to support fallback if the top action fails.
Second, it emits answer proposals: structured responses for current page questions, keyed to question identifiers and constrained by $\Gamma_t$.
Planning determines how the agent progresses through the interface, while answer generation determines what information is entered into the survey fields.

\textbf{(v) Execution and Replanning:}
An execution layer selects the highest-priority feasible action, invokes the corresponding tool, and applies the change to the webpage. 
The environment then transitions to a new state, yielding an updated context $C_{t+1}$.
The loop does not terminate if execution fails (e.g., stale element reference, blocked interaction). 
Instead, the interaction history incorporates failure signals, the page context refreshes, and replanning occurs under the updated state.
This adaptive replanning distinguishes the agent from deterministic automation scripts, enabling recovery from unforeseen interaction failures.

\textbf{(vi) Termination and Safeguards:}
The workflow terminates upon satisfying the completion criterion of detecting the terminal \texttt{/done} page. 
Optional safeguards include maximum step limits, repeated-failure thresholds, and no-progress detection to prevent infinite loops.
Overall, the architecture enforces a clean functional decomposition: perception constructs state, the interface adapter defines admissible actions and answer constraints, the reasoning module selects navigation and answer strategies, and the execution layer applies actions to the environment.
\begin{table}[h]
    \centering
    \small
    \begin{tabular}{@{}l | p{.8\columnwidth}@{}}
    \textbf{Var} & \textbf{Meaning} \\\hline

    $t$ &
    Interaction timestep in the observe–plan–act loop \\\hline

    $C_t$ &
    Page context at step $t$ (DOM, text, metadata, screenshot) \\\hline

    $A_t$ &
    Valid action space derived from page elements \\\hline

    $\Gamma_t$ &
    Interaction constraints (e.g., required fields, formats) \\\hline

    $O_t$ &
    Structured agent output (action plan + answers) \\\hline

    \end{tabular}

    \vspace{0.5em}
    \(
    C_t \rightarrow (A_t, \Gamma_t) \rightarrow O_t \rightarrow C_{t+1}
    \)
    \normalsize
    \caption{\textbf{Control Loop Formalization:}
    At timestep $t$, the agent observes page context $C_t$, from which the interface adapter derives a valid action space $A_t$ and interaction constraints $\Gamma_t$. 
    The agent produces a structured output $O_t$ specifying an action and responses, which is executed by the environment to produce the next state $C_{t+1}$.
    }
    \label{tab:notation}
\end{table}

\section{Experiment 1: Exploitability Analysis}
\label{exp1}
\begin{figure*}[t]
    \centering
    \includegraphics[width=\textwidth]{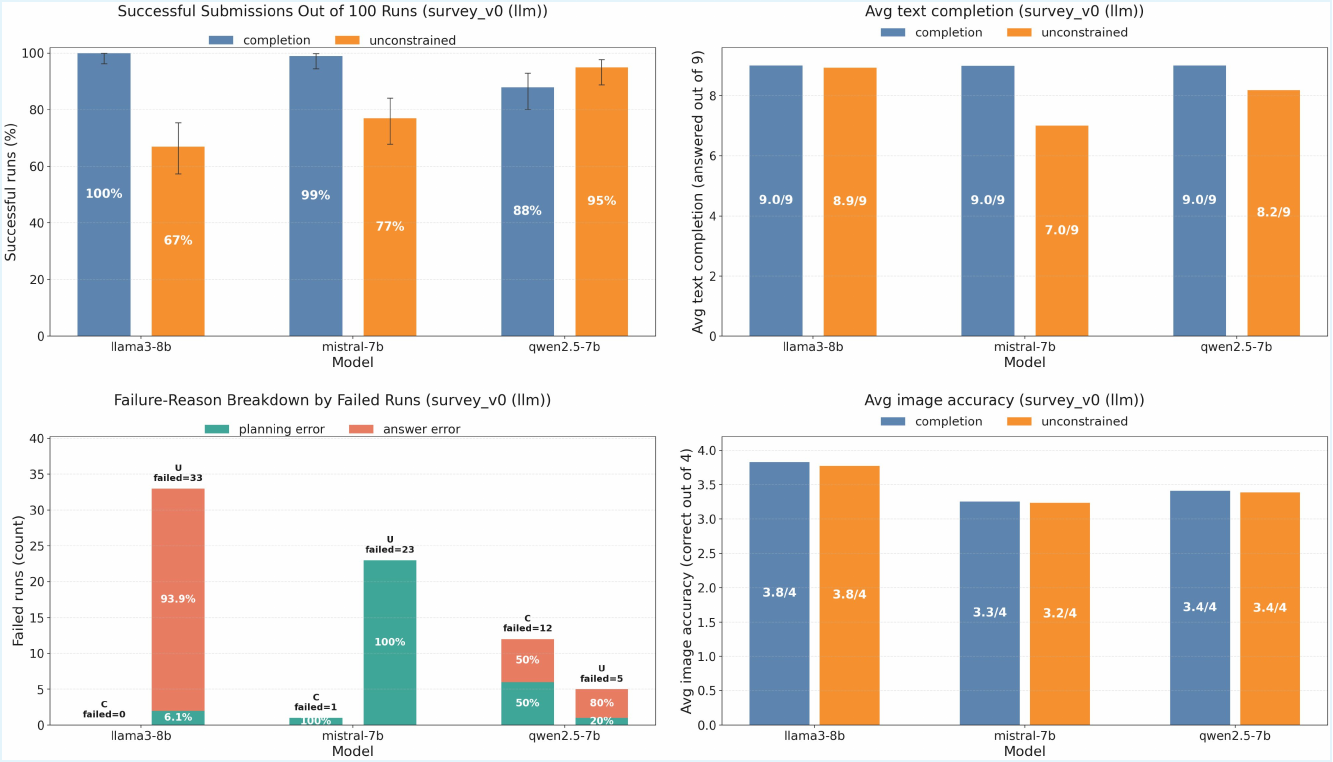}
    \caption{Experiment 1: Exploitability results on Survey\_v0. 
    Across models and behavioral policies, agents successfully complete the survey end-to-end while also achieving high page-level performance on both the text and image pages. 
    All models passed the embedded attention checks on both pages in 100\% of runs under both behavioral modes.
    }
    \label{fig:llm_v0_results}
\end{figure*}

We sought to evaluate how well an agent operates over structured DOM metadata and normalized browser interaction primitives to successfully complete the survey task.

\textbf{Models and Experimental Setup:}
We evaluate a set of open-source, instruction-tuned language models that are easy to deploy locally within our agent framework.
We deliberately use open-weight models to ensure reproducibility and experimental control. 
Local deployment is particularly important in our browser-interactive setting, which requires low-latency coordination between model outputs and tool execution; 
We conducted our experiments on a single machine with 20GB GPU memory and 32GB system RAM, which constrains feasible model sizes.
Under these constraints, we evaluate LLaMA-3-8B-Instruct, Mistral-7B-Instruct, and Qwen-2.5-7B-Instruct, hosted locally using LM Studio.
We evaluated all models under integration into the same agent pipeline without any architectural modifications, ensuring that performance differences arise from model capabilities alone.

\textbf{Behavioral Policies: Completion vs. Unconstrained}
We evaluate each model under two behavioral policies that govern how the agent can move through the survey.

\textit{(i) Completion Policy (Strict Mode):}
We add the following explicit instruction to the system prompt to simulate a fully compliant respondent who completes every field before proceeding:
\textit{“Hard requirement: do not click Next/Submit (or finish) until all answerable question items visible on the current page are answered.”}
This enforces a strict full-page \textit{answer-then-advance} strategy.
The agent must answer all visible answerable items on the page and perform exactly one final navigation click (Next/Submit) at the end of the plan.

\textit{(ii) Unconstrained Policy (Free Mode):}
Here, the hard requirement is not present.
The agent is free to advance without answering all visible items, skipping fields it chooses.
Therefore, lower completion rates in this mode are not necessarily indicative of failure. 

\textbf{Experimental Scale:}
Each of the 3 model evaluations occur in 2 behavioral modes, for 100 independent runs per mode.
This experiment required $\sim$3 hours of total compute time (3 models × 2 modes × 100 runs = 600 total agent runs). 

\subsection{Experiment 1: Results}

Figure~\ref{fig:llm_v0_results} summarizes performance across completion success, page-level accuracy, and failure decomposition.

\textbf{Overall Performance:}
Figure~\ref{fig:llm_v0_results} (Top-Left) reports end-to-end successful submissions, measured over 100 independent runs per model and behavioral mode. 
Across both completion and unconstrained policies, all models demonstrate strong end-to-end success. 
This indicates that the agent is capable of traversing the multi-page survey and reaching terminal submission states.

\textbf{Page-Level Performance:}
All page-level metrics are computed over successful runs (out of 100) for each model and mode.

\textit{(i) Text Page Completion.}
Figure~\ref{fig:llm_v0_results} (Top-Right) reports average text question completion (out of 9). 
In completion mode, text-page responses are nearly exhaustive across models, demonstrating that explicit prompt-level constraints effectively enforce full-page answer coverage. 
In unconstrained mode, answer coverage decreases for some models, indicating that agents sometimes move forward without answering all visible items. 
This confirms that the imposed behavioral policy strongly influences page-level completeness.

\textit{(ii) Image Page Accuracy:}
Figure~\ref{fig:llm_v0_results} (Bottom-Right) reports average image accuracy (out of 4).
Image-based question accuracy remains consistently high across both policies. 
The minimal gap between constrained and unconstrained settings indicates that visual attention checks are not inherently robust in this configuration. 

\textbf{Failure Mode Decomposition:}
Figure~\ref{fig:llm_v0_results} (Bottom-Left) reports the breakdown of failed runs out of 100 total executions per model and mode.
From the overall set of runs, we first determine how many did not result in successful end-to-end submission. 
Among those failed runs, we then categorize the primary failure cause as either a planning or answer error.
Planning errors occur when the agent fails to generate a valid next-step action (e.g., incorrect sequencing). 
Answer errors occur when the agent produces incorrect or invalid responses despite navigating the page successfully.
Across models, failures are distributed between these two categories, indicating that residual errors arise at the reasoning or answer-generation level rather than from instability in the interaction pipeline itself. 

\textbf{Structural Sources of Exploitability:}
The results demonstrate that modern instruction-tuned LLM agents are highly capable of completing the survey. 
However, beyond measuring capability, it is equally important to examine the reason for this success. 
While the agentic AI community view these results as evidence of robust autonomous form-completion, the survey and HCI communities must interpret them through a different lens: 
what aspects of standard web survey design unintentionally lower the barrier for autonomous completion?
In this subsection, we analyze the interface-level properties that enable reliable agent performance and discuss their implications for safeguarding online data collection.
Each survey page requires the agent to solve two subproblems. 
First is planning, namely determining the correct interaction primitive for each question (e.g., typing into a text input or clicking a button). 
Second is answer generation, which requires producing the appropriate response content when a question has a correct or expected answer. 
The extent to which these subproblems pose difficulty depends on the specific design of the page.

\textbf{Text-page:} ...presents no objectively correct or incorrect answers. 
The agent’s primary challenge is to determine the appropriate action type for each question.
Because the DOM is fully exposed, input types, required fields, value ranges, and selectable options are explicitly encoded in structured markup. 
The interface adapter further normalizes these elements into a valid action space. 
As a result, the agent does not need to infer interaction structure; it is directly provided with which primitive applies to which element.
In effect, the interface abstraction designed to simplify  browser interaction also lowers the reasoning burden. 
The agent can deterministically map each question to its correct interaction type using structured metadata alone. 
Since there are no right or wrong semantic answers on this page, exploitability reduces to correctly executing a sequence of structurally valid actions.

\textbf{Image-Page:} ...requires two steps for a respondent:
(1) visually reason about the images to determine the correct answer, and
(2) perform the correct interaction.
In contrast, the LLM-based agent does not rely on raw visual perception in this configuration. 
It operates over DOM and schema representations rather than pixel-level screenshots. 
In standard web design, images appear with metadata such as descriptive text, element attributes (e.g., \texttt{alt} tags), filenames, and structured identifiers. 
These attributes provide semantic cues about image content.
Consequently, what a survey researcher intended as a vision-based reasoning task is transformed into a language-based reasoning task for the agent.
The model receives textual descriptors and structured associations that provide additional context. 
The problem becomes similar to answering a question with enriched metadata rather than interpreting raw visual signals.
The same action-space simplification applies here: having inferred the correct answer from structured cues, selecting it is straightforward within the normalized interaction primitives.

\textbf{Design Trade-Off: Developer Transparency vs.\ Machine Robustness:}
The core source of exploitability lies in the semantic and structural transparency of the interface. 
Standard web survey design practices expose machine-readable cues that reduce ambiguity in both interaction and answer interpretation. 
This transparency is not accidental; 
it is a byproduct of modern web development practices that prioritize accessibility, maintainability, and developer efficiency. 
However, the same structured representations that simplify implementation and improve usability also provide rich semantic cues to autonomous agents. 
As agentic systems continue to advance, this developer-oriented clarity becomes a double-edged sword: 
it streamlines legitimate software development while simultaneously lowering the barrier for large-scale autonomous survey completion.
The central challenge, therefore, is not whether LLM agents \emph{can} complete surveys (they can), but how survey systems can balance usability, accessibility, and robustness against autonomous completion at scale.
\section{Experiment 2: Metadata Obfuscation}
\label{exp2}
\begin{figure*}[t]
    \centering
    \includegraphics[width=\textwidth]{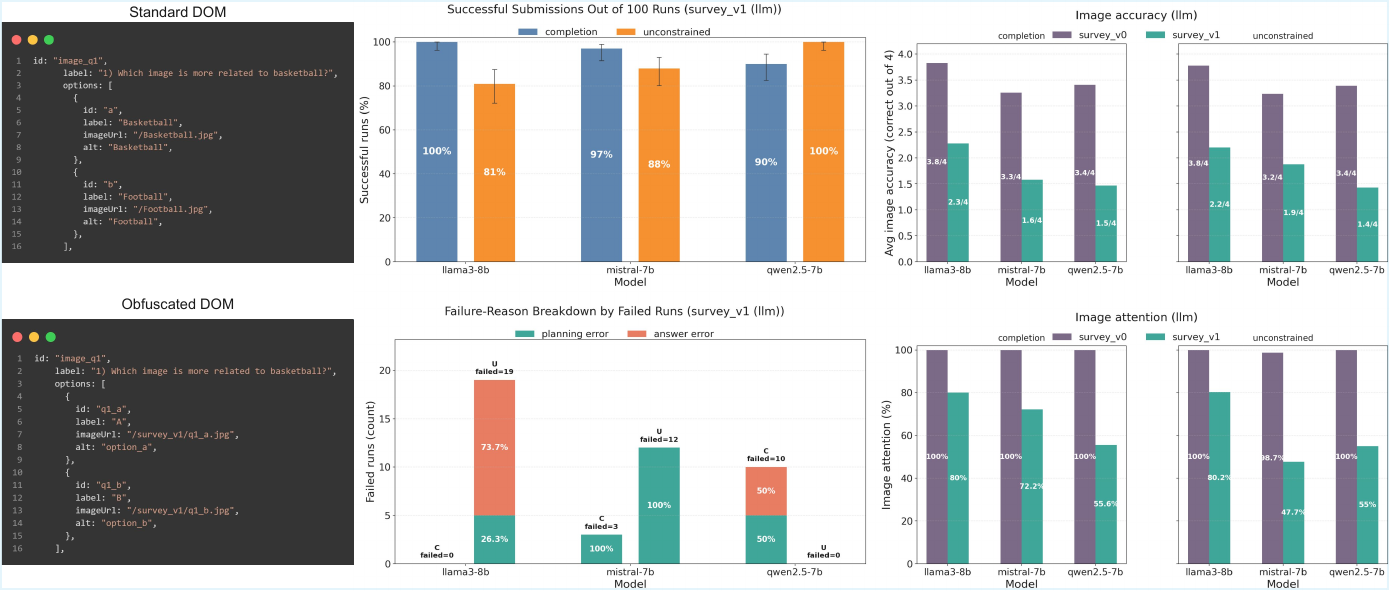}
    \caption{Results under metadata obfuscation (\texttt{survey\_v1}).
    Removing semantic DOM cues preserves workflow execution but significantly degrades image-answer accuracy and attention performance, isolating metadata exposure as a key driver of autonomous answer inference.
    }
    \label{fig:llm_v1_results}
\end{figure*}
In this section, we evaluate the extent to which we can mitigate the identified vulnerabilities through interface-level defensive modifications.
The analysis in Section~\ref{exp1} showed that the primary vulnerability lies in semantic cues embedded in the DOM—especially on the image page, where correct answers leak through the machine-readable metadata that expose information about image content. 
On the text page, questions are open-ended and have no objectively correct answers. 
While the DOM exposes structural information about input types and required fields, the absence of ground-truth correctness reduces risk of exploitation primarily to planning rather than answer inference. 
Moreover, such structural markup reflects standard web development practices and usability constraints.
For these reasons, our defensive strategy focuses on the image page, while the text page remains unchanged.

\textbf{Models and Experimental Setup:}
We construct a modified survey version, denoted \texttt{survey\_v1}, accessible via a separate URL endpoint. 
The core workflow remains identical to \texttt{survey\_v0}, but we modify the image page by replacing descriptive labels and alt-text with generic tokens, removing semantic meaning from filenames, and abstracting answer options into non-descriptive placeholders.
Figure~\ref{fig:llm_v1_results} (Left) illustrates the difference between the standard and obfuscated DOM representations.
Interaction primitives, page transitions, and overall survey structure remain unchanged, isolating semantic metadata as the sole manipulated variable.
We evaluate \texttt{survey\_v1} using the same agent pipeline, models, behavioral policies, and hardware configuration as in Section~\ref{exp1}.

\subsection{Experiment 2: Results}
Figure~\ref{fig:llm_v1_results} summarizes performance under the obfuscated configuration.
As in Section~\ref{exp1}, we compute page-level accuracy and attention metrics over successful end-to-end runs only. 
Per-page completion remains near 100\% across models and policies.

\textbf{End-to-End Completion:}
Figure~\ref{fig:llm_v1_results} (Middle-Top) shows that completion rates remain high across models and policies. 
Agents continue to navigate the multi-page workflow and reach terminal submission states, indicating that the defense does not disrupt functional survey completion.

\textbf{Image Page Accuracy:}
Figure~\ref{fig:llm_v1_results} (Right-Top) shows image accuracy drops substantially across all models and modes in \texttt{survey\_v1} compared to \texttt{survey\_v0}.
This indicates that exposed semantic metadata was a primary driver of correct answer inference in Experiment~\ref{exp1}. 
Once the survey obfuscates descriptive cues, models can no longer reliably align instructions with the correct image.
Agents continue to execute interface actions correctly, but fail at semantic reasoning about \textit{which option is correct}.

\textbf{Image Attention Check Performance:}
Figure~\ref{fig:llm_v1_results} (Right-Bottom) shows obfuscation drops attention accuracy substantially.
This confirms that the models previously relied on semantic DOM cues to satisfy visual attention checks. 
Without those cues, selection becomes closer to chance-level (50\% in all questions).

\textbf{Failure Mode Breakdown:}
Figure~\ref{fig:llm_v1_results} (Middle-Bottom) shows that errors in \texttt{survey\_v1} shift primarily toward answer-generation failures rather than planning errors.
This is consistent with the design of the intervention:
Planning remains intact, but we disrupt semantic interpretation of correct answers.
The obfuscation specifically targets the answer-generation subproblem identified in Section~\ref{exp1}.

\textbf{Interpretation:}
Experiment 2 isolates semantic metadata as a key driver of autonomous answer accuracy. 
When we obfuscate machine-readable cues, performance on tasks with objectively correct outcomes degrades substantially, while navigation and interaction planning remain unaffected. 
Although completion remains possible, the reduction in answer accuracy suggests improved robustness of survey data quality under obfuscation.
This defense primarily targets semantic cues exposed through the DOM, motivating evaluation of models that operate over visual inputs.
\section{Experiment 3: Multimodal Evaluation}
\begin{figure*}[t]
    \centering
    \includegraphics[width=\textwidth]{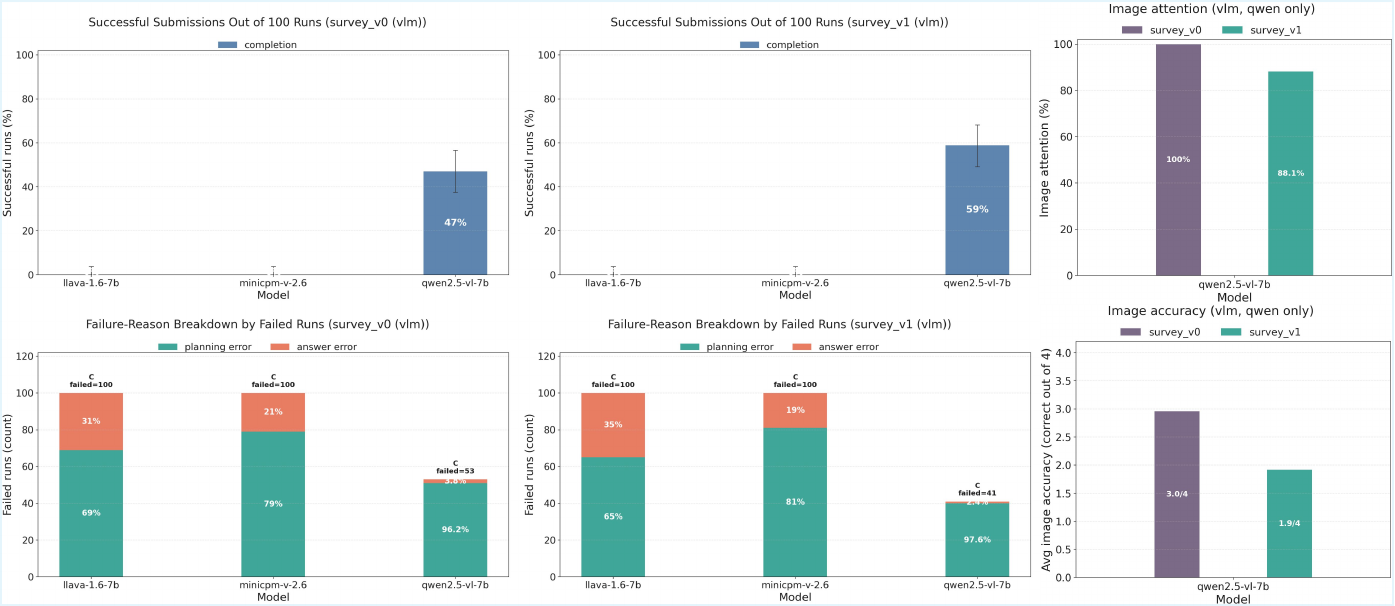}
    \caption{Results for multimodal agents across \texttt{survey\_v0} and \texttt{survey\_v1}. 
    While visual grounding partially preserves image-answer performance under metadata obfuscation, planning errors dominate overall VLM failures in both survey versions.
    }
    \label{fig:vlm_results}
\end{figure*}
After evaluating LLM agents on both survey versions, we next examine whether access to visual input alters performance.
The agent pipeline remains unchanged, the only modification is that, at each interaction step, the agent receives a rendered page screenshot in addition to the structured DOM representation.
This enables models with multimodal capability to ground reasoning directly in visual content rather than relying solely on machine-readable metadata.
Due to computational constraints and local deployment through LM Studio, we evaluate a limited set of open-weight VLMs that fit our compute budget.
Specifically, we test \texttt{llava-1.6-7b}, \texttt{minicpm-v-2.6}, and \texttt{qwen2.5-vl-7b}.
We conducted all VLM experiments in \emph{completion mode} only.

\textbf{VLM performance:}
Figure~\ref{fig:vlm_results} (Left and Middle) summarizes model performance across survey versions, demonstrating substantial variation. 
Only one model achieves moderate completion rates, while the others fail to produce successful end-to-end runs under the current pipeline configuration. 
Failure decomposition indicates that the dominant source of unsuccessful runs in both survey versions is planning error rather than answer-generation error. 
This contrasts with the text-only LLM results in Section~\ref{exp1} and~\ref{exp2}, where planning remained highly reliable across models.

\textbf{Image-Page Performance:}
For image-level metrics, we report results only for \texttt{qwen2.5-vl-7b}, as the other VLMs did not yield any successful end-to-end runs, preventing computation of per-page metrics.
As in previous experiments, per-page completion within successful runs remains near 100\%.
On \texttt{survey\_v0}, the Qwen VLM achieves near-perfect image attention performance and moderate image-answer accuracy. 
This indicates that when semantic DOM cues are available, multimodal grounding can effectively align visual content with task instructions.
On \texttt{survey\_v1}, attention performance remains high, suggesting that the model continues to correctly localize and inspect relevant visual elements. 
Image-answer accuracy decreases, but the degradation is substantially less severe than for text-only LLM agents~\ref{fig:llm_v1_results}.
This contrast highlights a key modality distinction: while text-only agents rely heavily on semantic DOM metadata for answer inference, VLMs can \textit{partially} compensate for its removal by grounding decisions directly in pixel-level visual content.

\section{Conclusion}

\textbf{For Agentic AI Researchers:}
Our results suggest that autonomous agent failures stem from both planning instability and answer-generation errors, rather than a single failure mode. 
This raises broader questions about how perception and structured planning interact: 
whether visual grounding increases representational complexity, how multimodal state encoding affects action selection, and why closed-loop agents fail under sequential tasks. 
Future work should examine these issues through systematic error analysis to better understand agentic system behavior.

\textbf{For Survey Researchers:}
We offer two recommendations to reduce data leakage: maintain separate answer keys and limit information exposed through the DOM. 
Our results show that language-only models can solve vision-based tasks when answers are leaked through DOM metadata. 
However, as VLMs improve, current defenses may become less reliable, making it increasingly difficult to distinguish human respondents from capable agents. 
This raises the question of whether some surveys should reconsider in-person methods despite their cost. 
More broadly, our findings point to an evolving arms race: survey designers must adapt to increasingly capable agents, while avoiding defenses that overly burden or exclude legitimate participants.

\section*{Impact Statement}
This paper presents work whose goal is to advance the field of Machine
Learning. There are many potential societal consequences of our work, none
which we feel must be specifically highlighted here.







\bibliography{main}
\bibliographystyle{icml2026}

\newpage
\appendix
\onecolumn
\clearpage

\section{Code Availability} 
All agent and survey files are available on GitHub:

\begin{center}
\url{https://github.com/SouravPanda11/Attention-Agents}
\end{center}
The repository includes a detailed README with setup instructions.

\section{Related Work}


\textbf{Attention checks for online survey:}
Attention check mechanisms are widely used to improve response quality in online data collection.
These mechanisms are designed to verify that respondents are actively reading and following instructions, thereby reducing inattentive or careless responses. 
In addition, attention checks function as a safeguard against automated or non-human submissions (e.g., bots), serving as a first line of defense for preserving data integrity. 
The design, effectiveness, and limitations of attention checks remain a recurring topic of research interest.
Abbey and Meloy \cite{Abbey2017} examined three forms of attention checks—directed queries, logical consistency statements, and manipulation checks—and found that combining multiple types increases the likelihood of detecting inattentive respondents.
Similarly, Kung et al. \cite{Kung2018} showed that instructional-manipulation and instructional-response checks can be implemented without compromising survey scalability. 
More recently, Roth and Yakobi \cite{Roth2024} provided empirical validation that even simple attention checks can effectively identify inattentive participants under conventional human-response settings.
While these studies demonstrate the utility of attention checks for filtering careless human responses, their robustness is increasingly challenged by advances in automated systems. 
\cite{Pei2020} showed that machine learning models can be trained to defeat traditional attention-check mechanisms. 

Likewise, Westwood \cite{westwood2025potential} demonstrated that AI agents are capable of successfully passing embedded attention checks, raising concerns about the continued reliability of these safeguards.
Taken together, this emerging body of work suggests that attention checks were originally designed under a threat model centered on inattentive humans or simple automation. 
The rise of more capable AI systems introduces qualitatively different risks. 
Rather than merely exposing vulnerabilities, these findings underscore the need for rethinking data-quality safeguards in online surveys to ensure continued research reliability.

\textbf{Agentic AI simulating End-users' capabilities:}
Recent advances in agentic AI have enabled systems that perform digital tasks in ways that closely resemble human behavior. 
These systems integrate large language models with tool-use capabilities, allowing them to perceive, reason about, and act within web environments. 
Commercially deployed agents such as Manus \cite{manus2025browseroperator} and Operator \cite{openai2025operator} demonstrate the ability to autonomously navigate websites, fill forms, and complete multi-step workflows—for example, browsing e-commerce platforms and executing purchases without manual intervention.
Beyond task completion, such agents can simulate end-user behavior across a range of digital contexts. 
In search environments, USimAgent \cite{zhang2024usimagent} emulates human search behavior by generating queries, selecting results, and determining stopping points based on behavioral modeling. 
Other work shows that synthetic agents can be parameterized to reflect specific demographic attributes or political preferences and then used to influence polling estimates with only a small number of injected responses \cite{harley2025fakeSurveyAI}.

Westwood (2025) builds on this idea by showing that a fairly simple autonomous synthetic respondent can get around 99.8\% of standard attention checks and systematically skew online survey estimates~\cite{westwood2025potential}. 
CloudResearch and others have confirmed that modern AI agents easily pass normal attention checks as well as typical participants~\cite{cloudresearch2025aiagentdetection}.
In short, these AI systems can behave just like a survey respondent, and they read questions, choose answers, and produce text in a coherent way.

Under the hood, agents use the webpage’s underlying structure to answer questions.
They often rely on the page’s code and metadata (like HTML labels or ARIA accessibility tags) rather than visual cues.
Prototype bots linked to LLMs can understand diverse question formats, answer both closed and open-ended questions, and handle multiple questions per page, including basic attention checks and some types of CAPTCHAs~\cite{hohne2025bots}.

\section{Full-page Screenshots}
\label{appendix:full-page}
\begin{figure*}[t]
    \centering
    
    \begin{subfigure}[t]{0.48\linewidth}
        \centering
        \includegraphics[width=\linewidth]{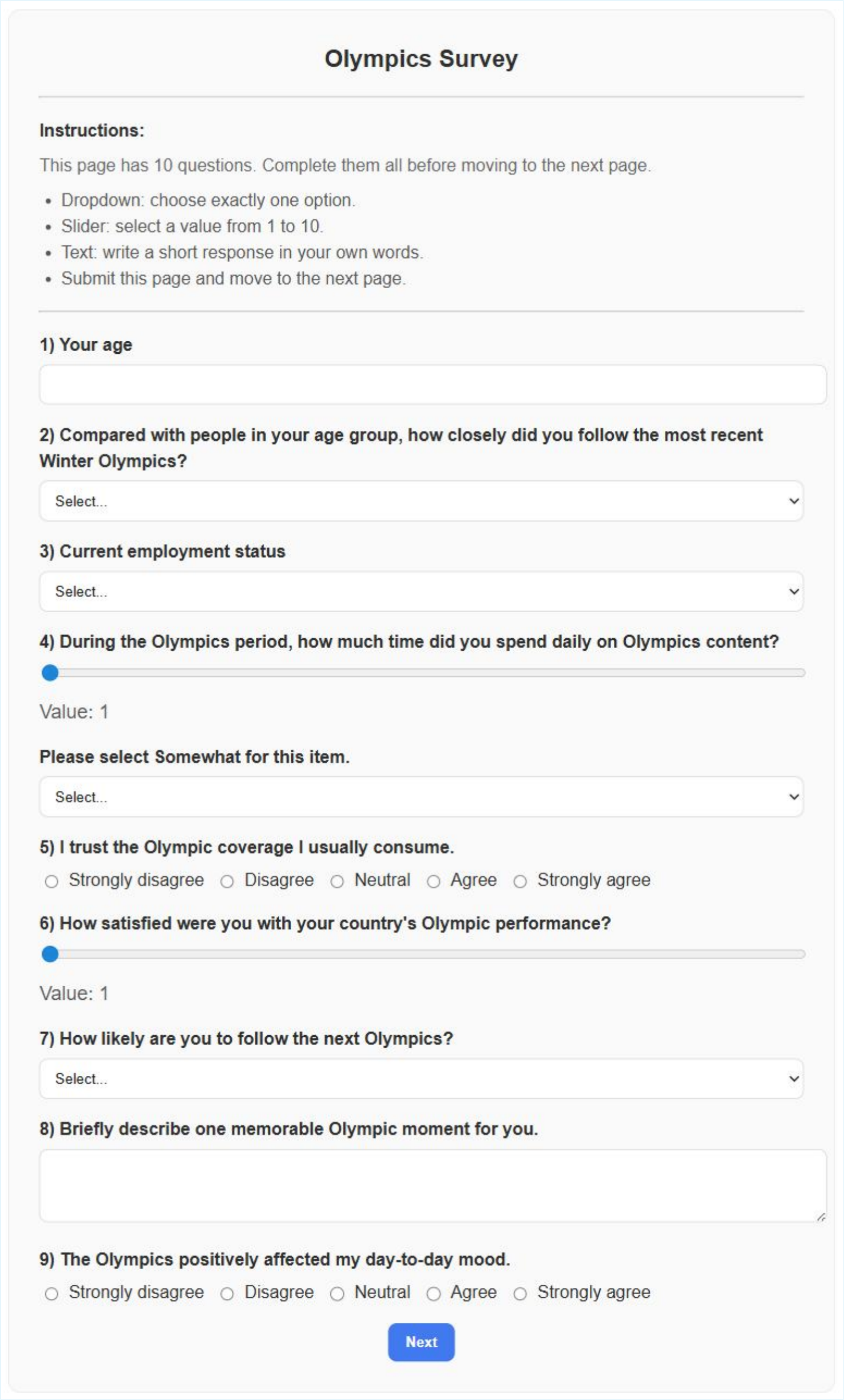}
        \caption{Text Page}
        \label{fig:text_page}
    \end{subfigure}
    \hfill
    \begin{subfigure}[t]{0.48\linewidth}
        \centering
        \includegraphics[width=\linewidth]{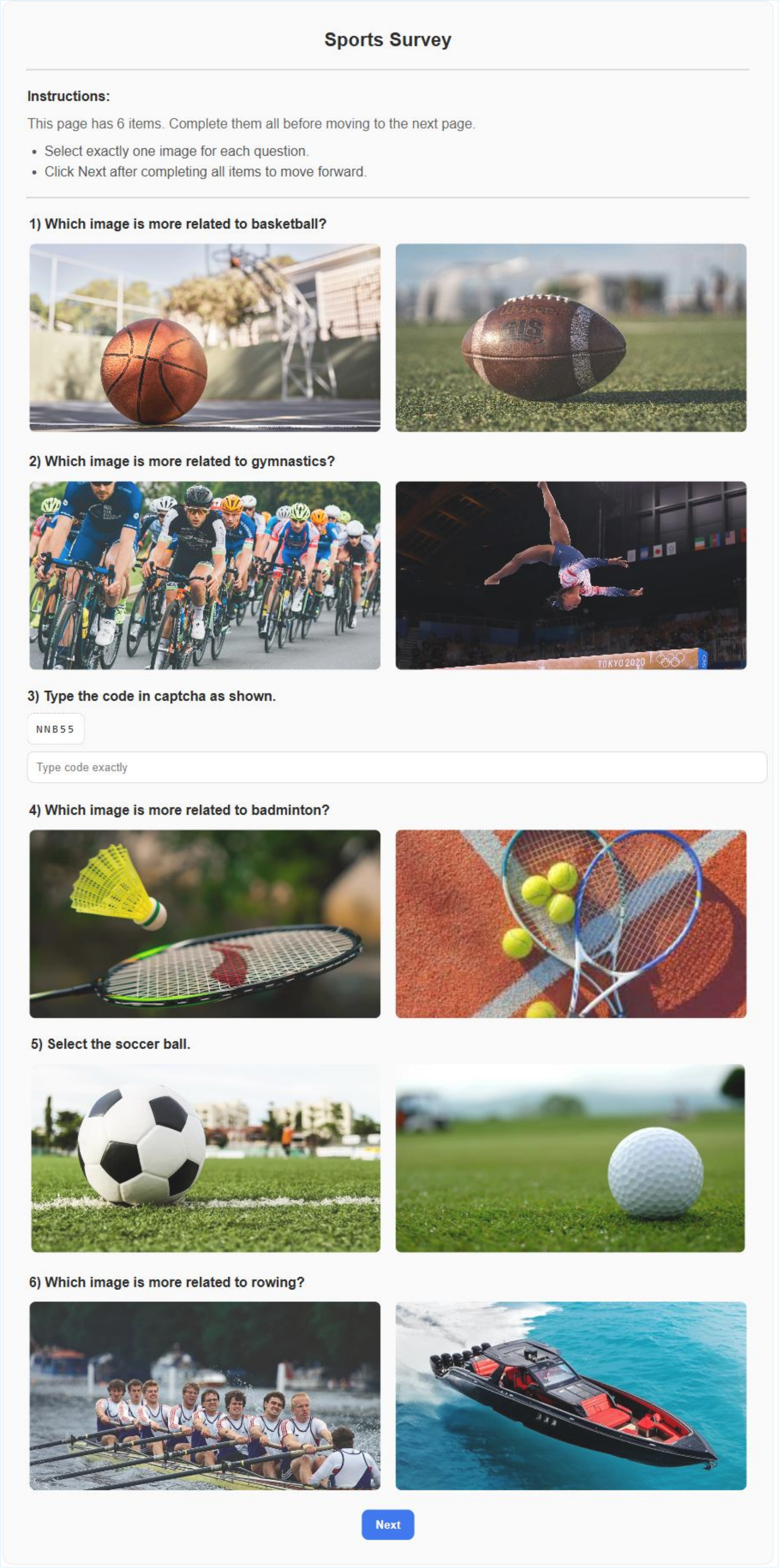}
        \caption{Image Page}
        \label{fig:image_page}
    \end{subfigure}
    
    \vspace{0.5em}
    
    \begin{subfigure}[t]{0.48\linewidth}
        \centering
        \includegraphics[width=\linewidth]{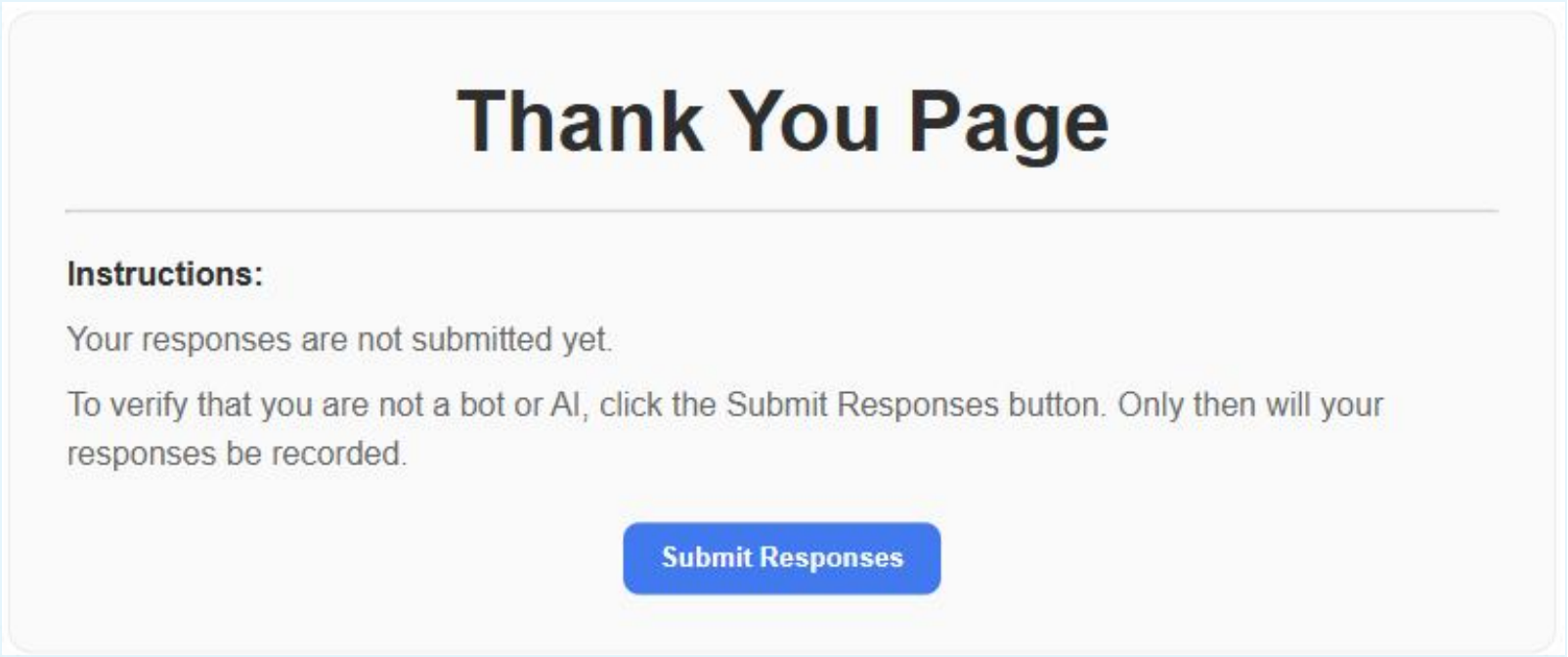}
        \caption{Thank You Page}
        \label{fig:thankYou_page}
    \end{subfigure}
    \hfill
    \begin{subfigure}[t]{0.48\linewidth}
        \centering
        \includegraphics[width=\linewidth]{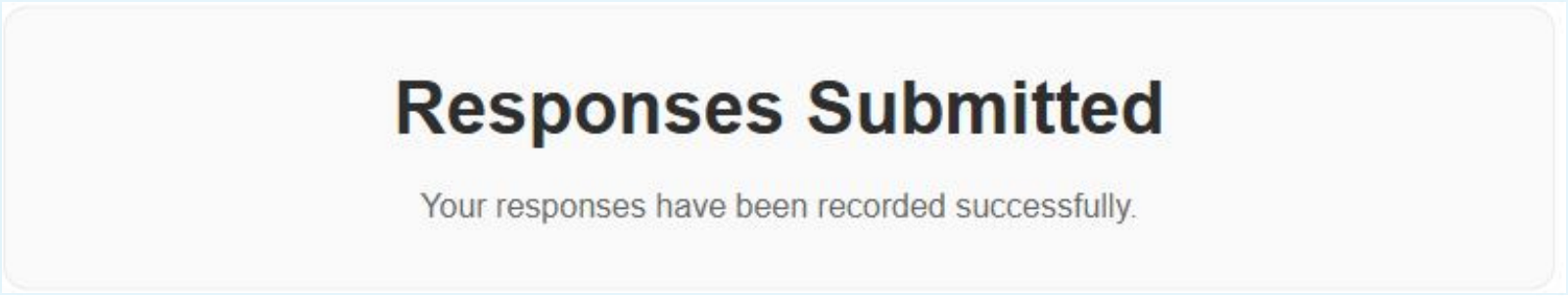}
        \caption{Submitted}
        \label{fig:done_page}
    \end{subfigure}
    
    \caption{Survey Sandbox}
    \label{fig:survey_screenshot}
\end{figure*}

Figure~\ref{fig:survey_screenshot} shows the full-page screenshots.

\end{document}